\documentclass[runningheads]{llncs}
\usepackage{graphicx}
\usepackage{amsmath}
\usepackage{amsfonts}
\usepackage{mathrsfs}
\usepackage{booktabs}
\usepackage{tikz}
\newcommand*\circled[1]{\tikz[baseline=(char.base)]{
            \node[shape=circle,draw,inner sep=0.3pt] (char) {#1};}}

\newcommand{\hyperbrain}{\textsc{HyperBrain}}
\newcommand{\setmixer}{\textsc{SetMixer}}
\newcommand{\brainwalk}{\textsc{BrainWalk}}
\newcommand{\setwalk}{\textsc{SetWalk}}

\newcommand{\mlpmixer}{\textsc{MLP-Mixer}}
\newcommand{\V}[0]{\mathcal{V}}
\newcommand{\E}[0]{\mathcal{E}}
\newcommand{\GG}[0]{\mathcal{G}}
\newcommand{\R}[0]{\mathbb{R}}

\newcommand{\bw}[0]{\textsc{Bw}}

\newtheorem{dfn}{Definition}

\begin{document}
\title{\hyperbrain{}: Anomaly Detection for Temporal Hypergraph Brain Networks}
%
%
\author{Sadaf Sadeghian \inst{1} 
\and Xiaoxiao Li \inst{2,3} 
\and Margo Seltzer \inst{1}
}
\authorrunning{Sadeghian et al.}
%
\institute{Department of Computer Science, University of British Columbia, Vancouver, Canada
\and
Department of Electrical and Computer Engineering, University of British Columbia, Vancouver, Canada 
\and
Vector Institute, Toronto, Canada\\
\email{\{sadafsdn, mseltzer\}@cs.ubc.ca }
}

\maketitle              
\begin{abstract}
    Identifying unusual brain activity is a crucial task in neuroscience research, as it aids in the early detection of brain disorders. It is common to represent brain networks as graphs, and researchers have developed various graph-based machine learning methods for analyzing them. However, the majority of existing graph learning tools for the brain face a combination of the following three key limitations. First, they focus only on pairwise correlations between regions of the brain, limiting their ability to capture synchronized activity among larger groups of regions.
    Second, they model the brain network as a static network, overlooking the temporal changes in the brain. Third, most are designed only for classifying brain networks as healthy or disordered, lacking the ability to identify abnormal brain activity patterns linked to biomarkers associated with disorders.
    To address these issues, we present \hyperbrain{}, an unsupervised anomaly detection framework for temporal hypergraph brain networks. 
    \hyperbrain{} models fMRI time series data as temporal hypergraphs capturing dynamic higher-order interactions. It then uses a novel customized temporal walk (\brainwalk{}) and neural encodings to detect abnormal co-activations among brain regions. 
    We evaluate the performance of \hyperbrain{} in both synthetic and real-world settings for Autism Spectrum Disorder and Attention Deficit Hyperactivity Disorder(ADHD).
    \hyperbrain{} outperforms all other baselines on detecting abnormal co-activations in brain networks. Furthermore, results obtained from  \hyperbrain{} are consistent with clinical research on these brain disorders.
    Our findings suggest that learning temporal and higher-order connections in the brain provides a promising approach to uncover intricate connectivity patterns in brain networks, offering improved diagnosis. Our code is available at: 
    \url{https://github.com/ubc-systopia/HyperBrain}.

\end{abstract}
\section{Introduction}
The brain is an intricate system, and functional magnetic resonance imaging (fMRI) is a widely-used neuroimaging technique for studying brain activity. 
Researchers often interpret fMRI data as a simple graph
with nodes representing regions of interest (ROI) and edges indicating functional connectivity through pairwise correlations of Blood-Oxygen-Level Dependent (BOLD) time series signals.
Recent advances in machine learning methods for analyzing graph-structured data have led to the development of effective approaches for studying human brain networks, particularly in tasks such as disease detection 
\cite{eslami2021machine,li2021braingnn}. These approaches classify brain states as healthy or indicative of a disorder. However, a crucial step in understanding symptoms and improving early detection of neurobiological disorders is identifying abnormal patterns in the brain. To address this gap, some studies 
\cite{li2021braingnn,abraham2017deriving}
 focus on brain network classification. They employ statistical tests or significant scores to pinpoint the most crucial regions or pairwise brain connections linked to identifying disorders. However, these approaches have drawbacks, such as depending heavily on classification accuracy, requiring a well-balanced dataset, which is rare in neuroimaging field and overlooking more complex patterns and structural features. 


Anomaly detection in the human brain is a promising solution for abnormal brain pattern discovery \cite{chatterjee2021detecting,caw,fan2021individual}, but it is a challenging task due to the lack of ground truth labeled anomalies and the need for a powerful brain modeling and analysis approach capable of capturing different patterns in the brain. 
Many existing anomaly detection methods are not designed for brain networks \cite{yadati2020nhp,caw,behrouz2024cat}. 
These methods can typically analyze only a single brain in isolation, making them inappropriate for fMRI due to the noisy data and the need of analyzing a group of brains to properly comprehend the disorder and capture dependable group-level biomarkers.
Some other methods designed for brain anomaly detection use non-learnable and fixed rules as anomalies,
which are not powerful and generalizable enough for the complex nature of brain activity and capturing patterns outside of the defined rules \cite{chatterjee2021detecting}.
Furthermore, most of these modeling approaches are limited in two different ways.


A limitation in previous fMRI based brain-modeling studies is that they often assumed that brain networks are static. 
However, recent research demonstrates dynamic changes in the brain\cite{chang2010time}, both in task-based fMRI \cite{gonzalez2018task} and resting-state fMRI 
\cite{hutchison2013dynamic},
revealing the dynamic nature and biologically meaningful evolution of brain activity.
Consequently, researchers have developed methods to track brain activity over time, including extracting dynamic functional connectivity or a temporal graph from fMRI time series 
\cite{hutchison2013dynamic,el2021dynamic} and using recurrent networks on the fMRI time series 
\cite{wang2019application}.
 By analyzing dynamic information, they can  improve detection accuracy of brain disorder.
However, leveraging dynamic information and temporal brain patterns in anomaly detection is under-explored.

Another modeling limitation in many previous brain analysis methods is their predominant focus on simple graphs \cite{li2021braingnn,fan2021individual,behrouz2023admire++}, ignoring the group activation of ROIs.
Clinical research indicates that cognitive mechanisms in the brain involve interactions among multiple co-activated brain regions, not just among pairs
\cite{lee2013resting}. 
Although others have improved brain classification accuracy using \emph{hypergraphs} to capture the complex relationships among ROIs by introducing hyperedges that connect multiple nodes simultaneously \cite{xiao2019multi,li2022construction,santoro2023higher},
they often neglect temporal patterns \cite{xiao2019multi,santoro2023higher,zu2016identifying} or limit the sizes of higher-order interactions, e.g., considering only interactions among three regions \cite{zu2016identifying}. Moreover, the crucial task of identifying abnormal patterns remains unaddressed \cite{li2022construction,santoro2023higher,zu2016identifying}.

We present \hyperbrain{}, a specialized framework for detecting abnormal co-activations in brain networks. 
\hyperbrain{} represents fMRI data as temporal hypergraphs, effectively capturing dynamic higher-order interactions in the brain. It then uses a novel temporal walk customized for brain networks, \brainwalk{}, to extract higher-order temporal motifs. Then, it learns the structural and temporal properties of brain networks through neural encodings for higher-order walks. Finally, \hyperbrain{} uses these encodings to calculate an anomaly score for each co-activation. 
By leveraging a training approach on diverse healthy brain networks, \hyperbrain{} enhances robust learning and mitigates noise, enabling it to identify anomalous hyperedges in the brains of individuals with disorders. Remarkably, \hyperbrain{} only relies on the neuroimaging data from healthy control group for training, eliminating the need for a balanced dataset of healthy and disordered subjects. 

\begin{figure}[t!]
  \centering
  \includegraphics[width=\linewidth]{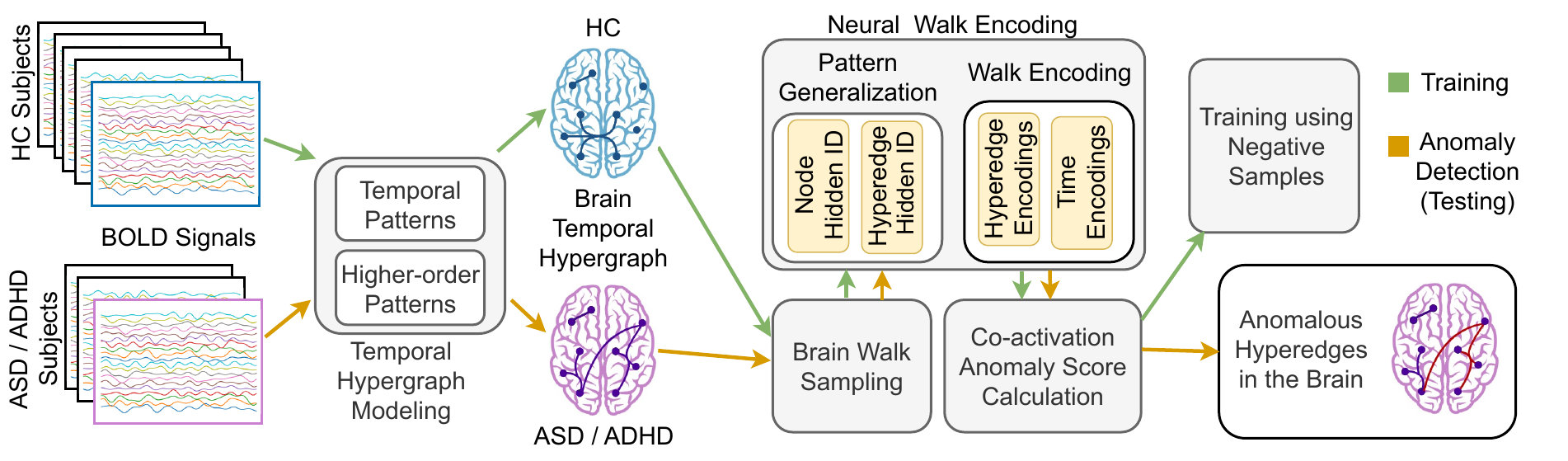}
  \caption{\textbf{Schematic of \hyperbrain{}}.  \hyperbrain{} consists of four stages: (1) Modelling both temporal and higher-order interactions among brain regions [{\S}\ref{sec:Modeling}], (2) Extracting temporal, higher-order patterns of brain activity  [{\S}\ref{sec:Brainwalk}] , (3)  Neural encoding to merge information from the sequence of hyperedges and their timestamps in the extracted brain patterns [{\S}\ref{sec:neuralWalkEncode}], and (4) Calculating anomaly scores for each brain co-activation [{\S}\ref{sec:anomalyScore}]. In training \hyperbrain{}, we only rely on healthy control data to detect anomalous co-activations in the brain [{\S}\ref{sec:hb_training}] .
  }
  \label{fig:framework_hb}
\end{figure}

Our experiments highlight \hyperbrain's outstanding performance in detecting abnormal brain co-activation associated with Attention Deficit Hyperactivity Disorder (ADHD) and Autism, outperforming all other baseline methods. 
Furthermore, our real-world experiments show \hyperbrain's ability to detect abnormal brain activity. Figure \ref{fig:framework_hb} illustrates the components of \hyperbrain{}.

\section{The Proposed Framework}
   \begin{dfn}[Temporal Hypergraph]\label{dfn:temp_hypergraph}
        A temporal hypergraph is defined as $\GG = (\V, \E)$, where $\V$ denotes the set of nodes, and $\E$ represents hyperedges occurring in the hypergraph over time. Specifically, $\E$ is defined as the set  $\E = \{ (e_1, t_1),(e_2, t_2),\dots\}$, where $e_i \in 2^\V $ represents a hyperedge, and $t_i$ denotes the timestamp when $e_i$ occurs. 
        
    \end{dfn}

    Our task is to detect anomalous hyperedges in the brain network. For each hyperedge in the brain temporal hypergraph $(e_{k}, t_{k}) \in \E$, we compute an anomaly score $ \varphi(e)$ which indicates the level of abnormality of the co-activation represented by $e_k$ at time $t_k$.

\subsection{Modeling fMRI Data as Temporal Hypergraph Brain Network} \label{sec:Modeling}
To capture both temporal and higher-order interactions among brain regions, we represent fMRI data as temporal hypergraphs. The set of Regions of Interest (ROIs), denoted by $\V = \{ v_1, \dots v_R \}$, is defined using brain parcellation atlases, with $R$ indicating the number of ROIs based on the atlas. Using the same atlas for all the subjects' fMRI data, the set of ROIs remains identical across individuals; $\GG_i = (\V, \E_{\GG_i})$ represents the temporal hypergraph of the $i^{th}$ subject.

To capture the dynamic patterns in brain activity, we model the temporal properties of brain networks using the sliding window technique 
\cite{peng2022gate}.
 For the BOLD signals of the $i^{th}$ subject $S_{i} \in \R ^{R \times T}$, with $T$ denoting the total fMRI time interval and $S_{i}[v_m]$ as the BOLD signal for the $m^{th}$ ROI, we divide $S_{i}$ into windows $ \{S_{i}^1, \dots, S_{i}^M \} $, where $M = \lfloor \frac{T-L}{s} \rfloor + 1$, $L$ is the window length, and $s$ is the stride between windows.


To capture higher-order connections and generate the hypergraph $\GG_i$, a subset of ROIs $ \{u_1 , \dots , u_k \} \in \V $ form a hyperedge if the similarities between their corresponding BOLD signals within a windows exceed a threshold. 
With a similarity measure function, $\zeta$, and $e = \{u_1 , \dots , u_k \}$, we have: 
\[
(e, t_p) \in \E_{\GG_i}   \textrm{ if }  \zeta( S_{i}^p[u_1], \dots , S_{i}^p[u_k]) \geq \textrm{Threshold}
\]
We calculate signal similarity using the Pearson correlation coefficient and form a hyperedge between a set of ROIs where each ROI is in the top $90^{th}$ percentile of positive correlations of all the other ROIs in the hyperedge. 


\subsection{Brain Walk Sampling} \label{sec:Brainwalk}
As a walk-based graph learning approach, we sample a set of random walks over our brain network to automatically extract temporal, higher-order patterns of brain activity.
To accommodate the unique characteristics of brain networks, we introduce \brainwalk{}. Inspired by \setwalk{} \cite{behrouz2024cat}, each \brainwalk{} consists of a random sequence of \emph{hyperedges}, allowing us to effectively capture the dynamics of higher-order brain networks. 
In contrast to many temporal networks, where timestamps represent discrete moments, in our representation, a timestamp represents a continuous interval (window) of BOLD signals, capturing distinct brain states. Considering this property, \brainwalk{} uses backward, timestamp-based  traversal to capture historical information and intra-timestamp traversal to capture patterns that co-occur within a span of time corresponding to brain activity.

A \brainwalk{} with length $\ell$ on $\GG= (\V, \E)$ is defined as:
\[
    Brain Walk : \:(e_1, t_{e_1}) \rightarrow (e_2, t_{e_2}) \rightarrow \dots \rightarrow (e_{\ell -1}, t_{e_{\ell -1}})\rightarrow (e_\ell, t_{e_\ell}),  
\]
where $e_i \in \E$, and consecutive pairs of $e_i$ and $e_{i + 1}$ represent neighboring hyperedges, satisfying the condition $t_{e_i} \geq t_{e_{i+1}}$. The notation $BW[i]$ represents the $i$-th pair in the walk, where $BW[i][0] = e_i$ and $BW[i][1] = t_{e_i}$. 

In our sampling approach, we take into account the temporal proximity of timestamps. This consideration is crucial for understanding the transitions between different states and the lasting effects of the previous task or state of the brain. Therefore, a closer timestamp is likely to be more relevant. To capture this temporal relevance, we use a biased sampling walk. We sample $(e, t)$, a neighboring hyperedge of a previously sampled hyperedge $(e_{prev}, t_{prev})$ with a probability that scales according to $\exp \left( \theta (t - t_{prev}) \right)$. Here, $\theta$ represents the hyperparameter for the sampling time bias.


\subsection{Neural Hyperedge Anomaly Detection}
\subsubsection{Neural Walk Encoding} \label{sec:neuralWalkEncode}
Research in graph learning has shown that anonymizing node identities enables models to perform well in an inductive setting and generalizing to unseen patterns by learning general rules unconstrained by specific node identities \cite{caw,behrouz2024cat}. Following them, to ensure model performance on unseen patterns, we use a two-step anonymization process to conceal hyperedge identities \cite{behrouz2024cat}. 
Initially, we anonymize node identities by replacing them with positional encodings, capturing the occurrence of nodes in different positions across a set of sampled \brainwalk{}s. 
Subsequently, to compute the anonymized encoding of a hyperedge, we aggregate the anonymized node identities corresponding to the nodes it connects. This aggregation is performed using \setmixer{} \cite{behrouz2024cat}, a permutation-invariant pooling strategy based on \mlpmixer{} \cite{mlp-mixer}.
Finally, we encode each \brainwalk{}. Specifically, during the encoding of a \brainwalk{}, $\hat{bw}$, we use \mlpmixer ~\cite{mlp-mixer} to merge information from the sequence of hyperedge encodings and their corresponding timestamp encodings, resulting in the calculation of $\text{ENC} (\hat{bw}) $. For encoding the hyperedge timestamps, we follow previous work on random Fourier features \cite{kazemi2019time2vec} to obtain a vector representation for each timestamp assigned to a hyperedge in the \brainwalk{}. 

\noindent\textbf{Anomaly Score} \label{sec:anomalyScore}
 To calculate anomaly scores for each hyperedge $e = \{u_1 , \dots , u_k \}$, we use a neural encoding module. This module processes a set of sampled \brainwalk{}s starting from each node within the hyperedge. The anomaly score, $\varphi (e) $, is computed as follows:
    \begin{equation}\label{eq:anomaly_score}
        \varphi(e) = 
        \texttt{MLP} 
        \Biggl( \Psi 
        \left( 
            \{ \mathbf{\text{ENC}_{u_1}} , \dots, \mathbf{\text{ENC}_{u_k}} \} 
        \right) 
        \Biggl)
    \end{equation}
    where
    $
         \text{ENC}_{u_i} = \frac{1}{N} \sum_{\hat{bw} \in \bw_{u_i}} \text{ENC} \left( \hat{bw} 
         \right)
    $.
    Here $\texttt{MLP}$ is a 2-layer perceptron, $\Psi$ is \setmixer{} \cite{behrouz2024cat} and $\bw_{u_i}$ is the set of $N$ sampled \brainwalk{}s. For each walk $\hat{bw} \in \bw_{u_i}$, it holds that $u_i \in \hat{bw}[0][0] \cap \hat{bw}[1][0]$.

\subsection{Training}\label{sec:hb_training}
In the training phase, to be able to learn individual-level 
as well as group-level patterns common among all subjects, we work with a set of healthy brains.
From each brain's fMRI data, we generate the corresponding hypergraph. 
Adopting the widely employed negative sampling approach \cite{chen2023survey}, we generate a negative sample for every hyperedge $e \in \E$ within the brain hypergraph. This involves keeping $50 \%$ of the nodes from ${e}$ and substituting the remainder with nodes from $\V - {e}$, resulting in a negative sample.
This approach could inadvertently produce a hyperedge, $e_j$, that is already present in the hypergraph while generating a negative sample for another hyperedge, $e_i$.
To overcome the limitation and ensure the reliability of these negative samples, we introduce a new step not found in prior work 
\cite{behrouz2024cat,caw}. We verify that any generated negative hyperedge does not appear in any timestamp of the healthy brain and has never been considered as a normal co-activation in the brain network.
Subsequently, we calculate the anomaly score, as defined in equation \ref{eq:anomaly_score}, for every hyperedge in the training set, including both normal hyperedges and negative samples. The framework is then trained using a contrastive learning approach. A key advantage of \hyperbrain{} is that it relies solely on data from the healthy control group for training, taking advantage of the abundance of healthy data.

To enhance learning across all subjects' brains in the training data and mitigate the impact of noise inherent to individual networks, we use a two-step approach: pre-training on a subset of healthy brain networks designated for training, followed by fine-tuning on the remaining brain datasets in the training data. This ensures that the model learns from all available brain networks in the training data, promoting a more comprehensive understanding of normal and anomalous patterns across diverse subjects.

\section{Evaluation}
Our evaluation addresses two questions about the performance of \hyperbrain{}:
\begin{enumerate}
\item How effectively does \hyperbrain{} perform in the task of anomalous hyperedge (abnormal co-activation) detection compared to the baselines? (Sec.~\ref{synthetic})
\item Are the abnormal activities detected by \hyperbrain{} in the brains of people with disorders meaningful and consistent with established research on the disorder's impact on the brain? (Sec.~\ref{real-world})
\end{enumerate}

\subsection{Datasets and Baselines}
We conducted experiments using two real-world fMRI datasets:
\circled{1} \emph{ADHD data} \cite{brown2012ucla} includes neuroimaging data from 50 subjects diagnosed with ADHD and 50 typically developing (TD) controls.
\circled{2} \emph{ASD data} \cite{craddock2013neuro} contains fMRI data from 45 individuals with Autism and 45 subjects in healthy control group. For brain parcellation, we used the CC200 \cite{craddock2012whole} atlas.

For baselines, we compare \hyperbrain{} with six state-of-the-art approaches in two group of methods: structured-based methods and embedding-based methods.
Structured-based Methods are: \circled{1} Common Neighbor (CN) \cite{newman2001clustering}: one of the most widely used metrics in anomaly detection based on graph structure. It quantifies overlap or similarity between sets of connections in a network, assuming more normal connections between nodes with a higher number of common neighbors. \circled{2} Jaccard Coefficient (JC) \cite{papadimitriou2010web}: a normalized version of CN. 
\circled{3} Adamic/Adar (AA) \cite{zhou2009predicting}: a weighted version of JC that assigns higher importance to less connected common neighbors.
Embedding-based Methods are: \circled{4} Principled Multilayer Network Embedding (PMNE) \cite{liu2017principled}: a multiplex graph learning method that analyzes functional connectivity by considering each subject as a different type of edge. \circled{5} Neural Hypergraph Link Prediction (NHP) \cite{yadati2020nhp}: a deep hypergraph learning method that analyzes static brain hypergraphs. \circled{6}  Causal Anonymous Walks (CAW-N) \cite{liu2021inductive}: a deep learning walk-based temporal graph learning method that analyzes dynamic functional connectivity. 


\subsection{Quantitative Evaluation on Synthetic Anomalous Hyperedges}\label{synthetic}

To assess \hyperbrain{}'s effectiveness in detecting anomalous hyperedges in the brain, we evaluate it on a set of control brain networks belonging to healthy subjects not used in the training phase. 
Synthetic anomalous hyperedges are injected into these networks following techniques from prior work \cite{chen2023survey}, with further enhancements explained in Sec.~\ref{sec:hb_training}. 
Subsequently, we deploy \hyperbrain{} to detect these synthetically injected anomalies, assessing performance using the Area Under the ROC Curve (AUC).



     \begin{table} [tpb!]
        \small
        \centering
        \caption{Performance in Anomalous Hyperedge Detection: Mean AUC (\%). The best result is indicated in boldface.}
        \label{tab:HEP-result}
        \hspace{-4ex}
        \resizebox{\linewidth}{!} {
        \setlength{\tabcolsep}{2pt}
        \def\arraystretch{2}
        \begin{tabular}{l c c c c c c r }
         \toprule
               & \multicolumn{3}{c}{Structured-based} & \multicolumn{4}{c}{Embedding-based} \\
        \cmidrule(lr){2-4}\cmidrule(lr){5-8}
          Datasets \textbackslash{} Methods &  CN\cite{newman2001clustering}  & JC\cite{papadimitriou2010web} & AA\cite{zhou2009predicting} & PMNE\cite{liu2017principled} & NHP\cite{yadati2020nhp} & CAW-N\cite{liu2021inductive}& \textbf{\textsc{\hyperbrain{}}}\\
        \midrule
         \midrule
          \textsc{ADHD} & $75.21 $ & $78.00$ & $75.47$ & $75.53$ & $72.12$ & $86.26$ & $\mathbf{92.33 }$\\
         \textsc{ASD} & $86.47 $ & $86.35$ & $86.66$ & $72.54$ & $82.42$ & $83.56$ & $\mathbf{93.78 }$\\
        \toprule
        \end{tabular}
        }
    \end{table}

    The results presented in Table \ref{tab:HEP-result} demonstrate that \hyperbrain{} outperforms the baselines by a large margin.
    Three key factors contribute to \hyperbrain{}'s superior performance: 
    \circled{1} capturing higher-order patterns, \circled{2} incorporating temporal changes in the brain, 
    and \circled{3} the exclusive design and training approach of \hyperbrain{} for brain network considering its unique properties.
    

\subsection{Clinical Relevance and Consistency on Real Datasets}\label{real-world}
    To answer the second question, mirroring real-world scenarios, we investigate anomalies detected by \hyperbrain{} in individuals with specific disorders. Our analysis focuses on assessing whether these detected anomalous hyperedges align with existing research findings related to these disorders.
    We train \hyperbrain{} on neuroimaging data from healthy brains and subsequently test the trained model on brains of individuals diagnosed with disorders. After detecting anomalous hyperedges using \hyperbrain{}, we record the occurrence of each region. Subsequently, we identify and report regions exhibiting statistically significant occurrences in the detected anomalous hyperedges.
    In the subsequent sections, we analyze identified regions linked to ADHD and ASD.

\begin{figure}[t!]
    \begin{minipage}{0.5\textwidth}
        \centering
      \includegraphics[width=0.97\textwidth]{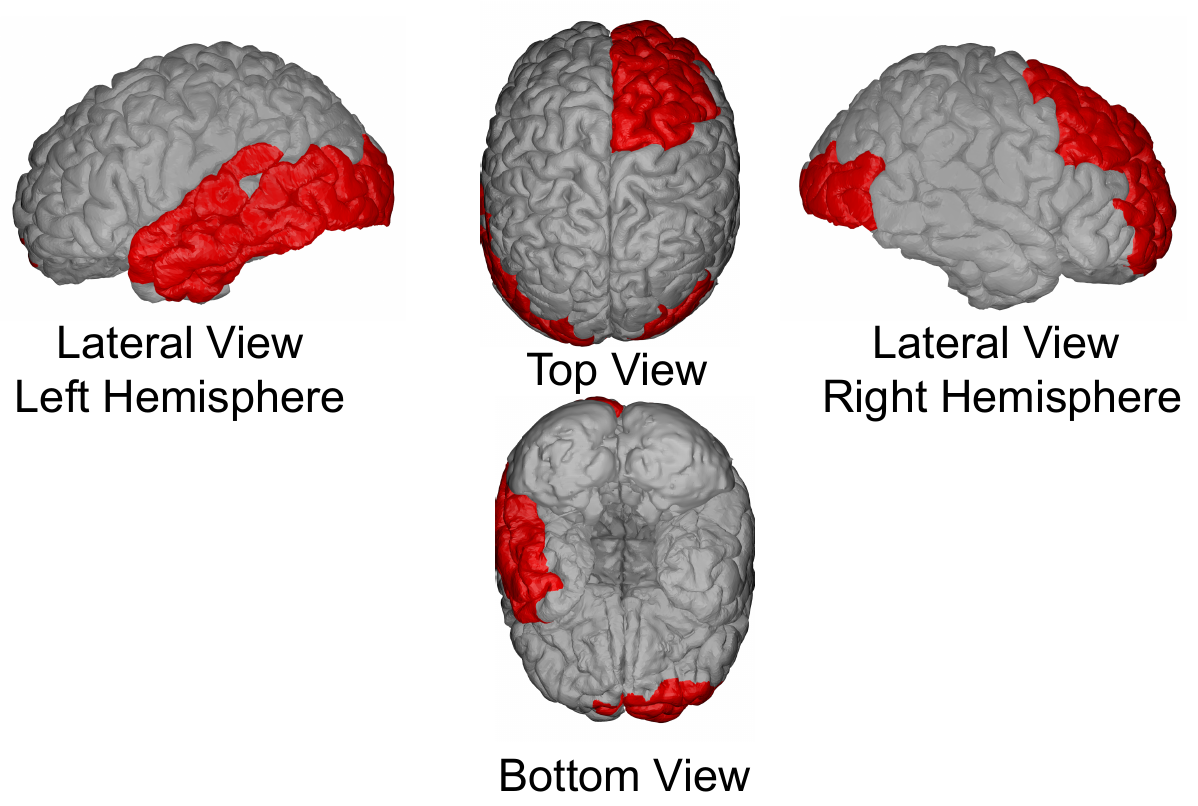} 
        \caption{ADHD-Related Brain Regions \\ Identified by \hyperbrain{}}
      \label{fig:adhd}
    \end{minipage}
    \begin{minipage}{0.5\textwidth}
        \centering
        \includegraphics[width=0.97\textwidth]{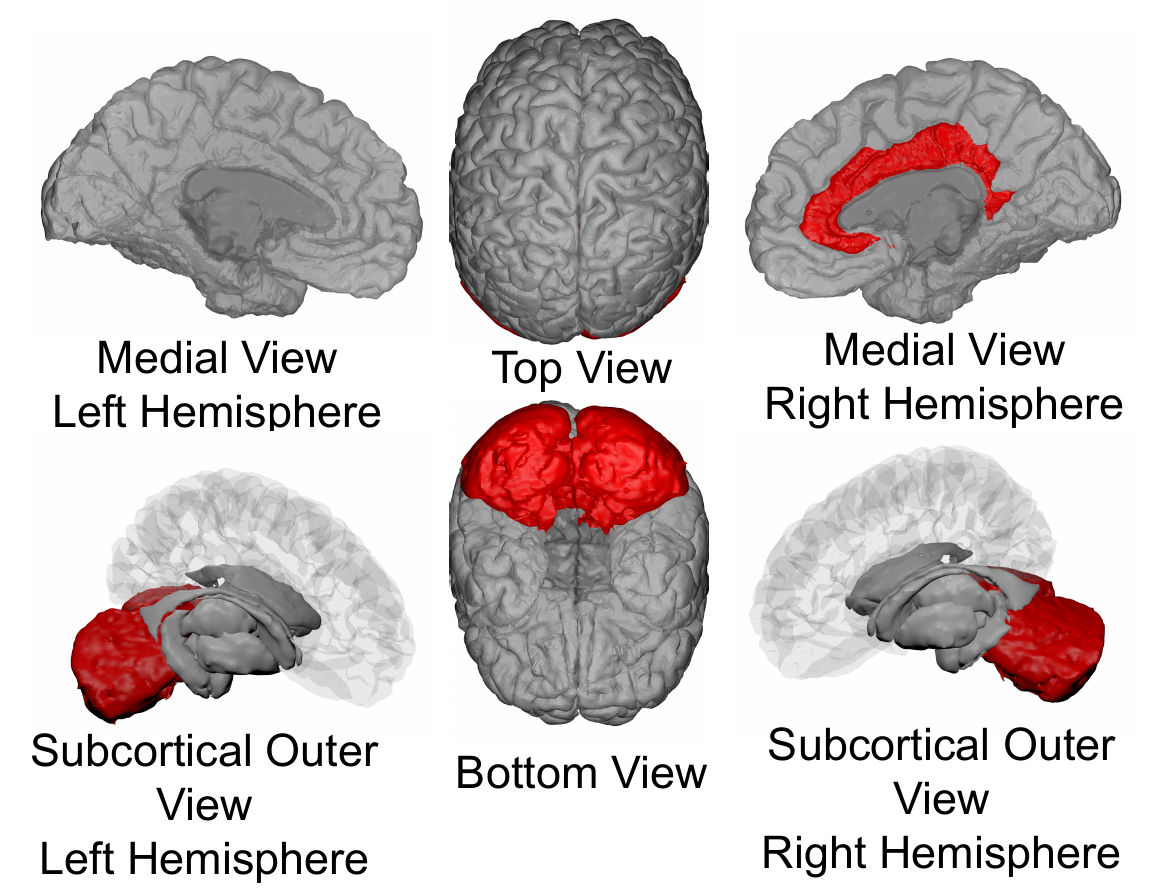}
        \caption{ASD-Related Brain Regions \\ Identified by \hyperbrain{}}\label{fig:asd}
    \end{minipage}
\end{figure}

    \noindent\textbf{ADHD}
    Brain regions with statistically significant occurrences in detected anomalous hyperedges include: \emph{Frontal Pole}, \emph{Right Frontal Gyrus}, \emph{Lateral Occipital Cortex}, and \emph{Left Temporal Gyrus} (refer to Figure \ref{fig:adhd}). Notably, the \emph{Frontal Pole} and \emph{Right Frontal Gyrus}, both located in the \emph{Prefrontal Cortex}, exhibit the highest occurrence rates. The \emph{Prefrontal Cortex} is known for its crucial role in attention regulation and has been associated with impaired function in individuals with ADHD 
    \cite{arnsten2009emerging}.
    The other anomalous regions identified in our study are also consistent with prior studies on ADHD
    \cite{soros2017inattention,yu2023meta}.
    Studies reported abnormally low activity in response to a stimulus in the \emph{Left Temporal Gyrus}\cite{yu2023meta}, as well as increased cortical thickness in the \emph{Occipital Cortex} \cite{soros2017inattention}, among individuals diagnosed with ADHD.

    
    \noindent\textbf{ASD}
    Brain regions statistically linked to these anomalous hyperedges include the
    \emph{Cerebellum} and \emph{Right Cingulate Gyrus} (refer to Figure \ref{fig:asd}). Notably, the \emph{Cerebellum}, the most frequently occurring region in detected abnormal hyperedges
    has been recognized as a key brain area affected in autism 
    \cite{rogers2013autism}.
    Additionally, the other detected region, the \emph{Right Cingulate Gyrus}, has been reported to exhibit atypical structure and activity in autistic brains \cite{chien2021altered}.

\section{Conclusion}

We introduce \hyperbrain{}, a novel method to identify biomarkers associated with neuro disorders via identifying anomaly patterns of high-order interactions among brain regions. To this end, we model fMRI data as temporal hypergraphs to effectively capture dynamic higher-order interactions. \hyperbrain{} uses higher-order random walks and a neural encoding to learn intricate patterns in the network. Trained on a set of healthy brain networks, it identifies anomalous patterns in the brain of individuals with disorders. Our evaluation shows that \circled{1} \hyperbrain{} outperforms other baselines in hyperedge anomaly detection, and \circled{2} the detected abnormal brain activities align consistently with clinical research on disorders. 
These results suggest several promising directions for future research, including deeper exploration of brain networks, enhanced understanding of symptoms, and early disorder detection. Additionally, as computational cost is a common challenge in hypergraph analysis, another direction for future work is improving the computational efficiency of our method, where the cost is $O(\#subjects \times \#temporal\_hyperedges)$.

\section{Acknowledgments and Disclosure of Funding}
We acknowledge the support of the Natural Sciences and Engineering Research Council of Canada (NSERC).\\
Nous remercions le Conseil de recherches en sciences naturelles et en génie du Canada (CRSNG) de son soutien.

\bibliographystyle{splncs04}
\bibliography{paper}
%






\end{document}